# Error Concealment by Means of Motion Refinement and Regularized Bregman Divergence

Alessandra Martins Coelho[1], Vania V. Estrela[2], Felipe P. do Carmo[3] and Sandro R. Fernandes[4]

[1]Instituto Federal de Educacao, Ciencia e Tecnologia do Sudeste de Minas Gerais (IF SEMG),Rio Pomba, MG, Brazil
`alessandra.coelho@ifsudestemg.edu.br`
[2]Telecommunications Department, Universidade Federal Fluminense (UFF), CEP 24210-240, Niterói, RJ, Brazil
`vestrela@id.uff.br`
[3]Colégio Pedro II, Rua Bernardo de Vasconcelos, 941, Realengo, CEP 21710-261, Rio de Janeiro, RJ, Brazil
`felcarmo@yahoo.com.br`
[4]Instituto Federal de Educação, Ciência e Tecnologia do Sudeste de Minas Gerais. Rua Bernardo Mascarenhas, 1283 Fábrica CEP 36080-001, Juiz de Fora, MG, Brazil
`sandro.fernandes@gmail.com`

**Abstract.** This work addresses the problem of error concealment in video transmission systems over noisy channels employing Bregman divergences along with regularization. Error concealment intends to improve the effects of disturbances at the reception due to bit-errors or cell loss in packet networks. Bregman regularization gives accurate answers after just some iterations with fast convergence, better accuracy and stability. This technique has an adaptive nature: the regularization functional is updated according to Bregman functions that change from iteration to iteration according to the nature of the neighborhood under study at iteration *n*. Numerical experiments show that high-quality regularization parameter estimates can be obtained. The convergence is sped up while turning the regularization parameter estimation less empiric, and more automatic.

## 1      Introduction

The extraction of motion information from a given video sequence is a major undertaking. In this *milieu*, optical flow (OF) (cf. [10]) is a recurrent concept with numerous applications, e.g., image compression/coding (cf. [11, 14]), automatic movie edition/digitalization (cf. [12]), reconstruction of 3D surfaces by means of depth from stereo (cf. [13]), object recognition and motion estimation (see, for example, [14-16]).

OF discontinuities are of particular interest and they should be distinguishable from object borders i.e., large gradients of grayscale values within the projections of moving objects. Furthermore, due to imperfect communication channels, it is nearly impossible to attain reconstructed pictures with suitable visual quality. That is why data protection and error reduction methods such as decoder-based error concealment (EC) algorithms are required. They rely on two types of redundancies: spatial (SEC) and temporal (TEC), requiring no alterations on the bit-stream syntax and transport technology. SEC algorithms: (a) interpolate the lost area using spatially neighboring image data; (b) presuppose statistical correlation between adjacent image blocks; and (c) provide a good approximation for the lost macroblocks (MBs). TEC schemes utilize previously decoded image data to estimate motion vectors (MVs) of the lost MBs to compensate for errors. EC is largely dependent upon the ability of the system to detect errors, since EC operations are applied to corrupted MBs. Because the damaged packetized bit-stream is thrown out and considered missing; we can obtain the MB position where an error occurs by checking the MB address (MBA), which defines the absolute position of the MB. The Bregmanized regularization relies on Bregman divergences constructed with the $q$-discrepancy functional [3, 4], so that the regularization function does not have to be fixed at each interaction [2, 5]. Information on the discrepancy principle can be used to improve the stopping criterion as well [3, 4]. The connection with exponential distributions allows for entropy-based estimation methods [8, 9] because numerous well-known divergences, such as relative entropy, can be expressed as Bregman divergences on the distribution parameters.

In this paper, the problem of error concealment in transmission of video over noisy channels is addressed. Section 2 states the motion estimation problem used in this text. Section 3 casts the problem in terms of the minimization of a regularization functional term depending on the Bregman divergence. The error concealment algorithm is introduced in Section 4. Experimental results are shown in Section 5. Finally, conclusions are drawn in Section 6.

## 2     The Motion Estimation Problem

OF is the distribution of apparent velocities of movement for intensities of pixels $I_k(r)$ of the $k$-th frame at location $r = [h, v]^T$ in an image and it requires at least two consecutive frames. The displacement of every pixel in a frame forms the displacement vector field (DVF). We seek the corresponding displacement vector (DV) $d(r) = [d_h, d_v]^T$ at the working point $r$, in the current frame $k$, in order to minimize the displaced frame difference (DFD) in an area containing the working point and assuming constant image intensity along the motion trajectory. The perfect registration of frames results in $I_k(r)=I_{k-1}(r\text{-}d(r))$. Then, the DFD can be written as $\Delta(r;d(r))=I_k(r)\text{-}I_{k-1}(r\text{-}d(r))$. An estimate of $d(r)$, is obtained via minimization of the gradient $\nabla I_{k-1}(r\text{-}d^i(r))$ or by determining a linear relationship between these two variables through some model. This can be accomplished by using a Taylor series expansion of $I_{k-1}(r\text{-}d(r))$ about the location $(r\text{-}d^i(r))$, where $d^i(r)$ represents a prediction of $d(r)$ in the $i$-th step. This results in, where the displacement update vector is $x=[x_h, x_v]^T = d(r) - d^i(r)$, and $e(r,$

$d(r)$) stands for the truncation error resulting from higher order terms and $\nabla=[\delta/\delta_h, \delta/\delta_v]^T$ represents the spatial gradient operator [17]. Considering all points in a neighborhood of pixels around $r$ leads to

$$y = Hx + \eta, \qquad (1)$$

where the temporal gradients $\nabla I_{k-1}(r-d^i(r))$ have been stacked to form $y$ containing DFD information on the pixels in a neighborhood, $H$ contains the spatial gradient operators at each observation, and the error terms have formed the additive white noise vector $\eta$. the corrected DV is given by

$$d^{i+1}(r) = d^i(r) + x^i(r), \qquad (2)$$

at iteration $i$. The ordinary least squares (LS or OLS) estimate of the update vector is

$$\hat{x}_{LS} = (H^T H)^{-1} H^T y. \qquad (3)$$

## 3  The Motion Recovery Algorithm

Segmenting OF via EM algorithm for mixtures of DVs can be done successfully [15] because it is presumed that there is little or no interference amid individual sample constituents or that all the constituents in the samples are known ahead of time. The Tikhonov regularization functional (TRF) (cf. [1, 2]) associated to Eq. (1) is

$$Q(\hat{x}) = \sum_{i=0}^{M-1}\sum_{j=0}^{M-1}\left[y(i,j) - \sum_{k=-N}^{N}\sum_{l=-N}^{N} b(k,l)\hat{x}(i+k,j+l)\right]^2 \\ + \alpha \frac{1}{1+q} \sum_{i=0}^{M-1}\sum_{j=0}^{M-1}\left\{\hat{x}_{i,j}\left[\frac{(\hat{x}_{i,j})^q - (\overline{x}_{i,j})^q}{q}\right] - (\overline{x}_{i,j})^q(\hat{x}_{i,j} - \overline{x}_{i,j})\right\}, \qquad (4)$$

where $\alpha$ is the regularization parameter, and is the estimate of $x$ obtained with Eq. (4). $S$ corresponds to a family of functions (Bregman divergences) given by

$$S = D_q(\hat{x}, \overline{x}) = \frac{1}{1+q}\sum_{i=0}^{M-1}\sum_{j=0}^{M-1}\left\{\hat{x}_p\left[\frac{(\hat{x}_{i,j})^q - (\overline{x}_{i,j})^q}{q}\right] - (\overline{x}_{i,j})^q(\hat{x}_{i,j} - \overline{x}_{i,j})\right\}, \qquad (5)$$

$\overline{x}$ stands for a reference value and $q$ is an adjustable parameter ($q$-discrepancy). The nonlinear system becomes

$$F_{rs}(\hat{x}) = \frac{\partial Q(\hat{x})}{\partial \hat{x}_{rs}} = 0, \qquad (6)$$

for $r,s = 0,1,2,...,M-1$. We seek the $q$ and $\hat{x}^t$ that minimize Eq. (6), where $t$ is an iteration counter and $\hat{x}^0$ is the initial estimate of $x$. The Newton-Raphson method yields

$$\hat{x}^{t+1} = \hat{x}^t + \Delta\hat{x}^t, \quad t = 0,1,2,..., \tag{7}$$

A first-order Taylor expansion of Eq. (6) results in

$$F_{rs}(\hat{x}^{t+1}) = F_{rs}(\hat{x}^t + \Delta\hat{x}^t) = F_{rs}(\hat{x}^t) + \sum_{m=1}^{M-1}\sum_{n=1}^{M-1} \left.\frac{\partial F_{rs}}{\partial \hat{x}_{mn}}\right|_{\hat{x}^t} \Delta\hat{x}^t_{mn} = 0. \tag{8}$$

With the help of the Gauss-Seidel method, we find

$$\Delta\hat{x}_{rs}\bigg|^{t,c+1} = -\frac{1}{\left(\partial F_{rs}/\partial \hat{x}_{mn}\right)\bigg|^{t,c}_{\substack{m=r\\n=s}}} \left\{ F_{rs}\big|_{\hat{x}^{t,c}} + \sum_{\substack{m=0\\m\neq r}}^{M-1}\sum_{\substack{n=0\\n\neq s}}^{M-1} \left.\frac{\partial F_{rs}}{\partial \hat{x}_{mn}}\right|_{\hat{x}^{t,\tilde{c}}} \cdot \Delta\hat{x}^{t,\tilde{c}}_{mn} \right\}, \tag{9}$$

with $\Delta\hat{x}^{t,0} = 0$, $c$ is the iteration counter and

$$\tilde{c} = \begin{cases} c+1, & \text{if } (m<r) \text{ or } (m=r \text{ and } n<s) \\ c, & \text{otherwise} \end{cases}. \tag{10}$$

Now, the update term becomes

$$\Delta\hat{x}_{rs}\bigg|^{t,c+1} = \Delta\hat{x}_{rs}\bigg|^{t,c} - \frac{1}{\left(\partial F_{rs}/\partial \hat{x}_{mn}\right)\bigg|^{t,c}_{\substack{m=r\\n=s}}} \left\{ F_{rs}\big|_{\hat{x}^{t,c}} + \sum_{\substack{m=0\\m\neq r}}^{M-1}\sum_{\substack{n=0\\n\neq s}}^{M-1} \left.\frac{\partial F_{rs}}{\partial \hat{x}_{mn}}\right|_{\hat{x}^{t,\tilde{c}}} \cdot \Delta\hat{x}^{t,\tilde{c}}_{mn} \right\}, \tag{11}$$

Once the corrections $\Delta\hat{x}_{RS}$ are calculated, new estimates $\Delta\hat{x}^{t+1}$ can be obtained from Eq. (11) and with a convergence factor $\gamma$, where $0 < \gamma < 1$, Eq. (7) becomes

$$\hat{x}^{t+1} = \hat{x}^t + \gamma \Delta\hat{x}^t, \quad t = 0,1,2,.... \tag{12}$$

The previous expression converges faster than Eq. (7).

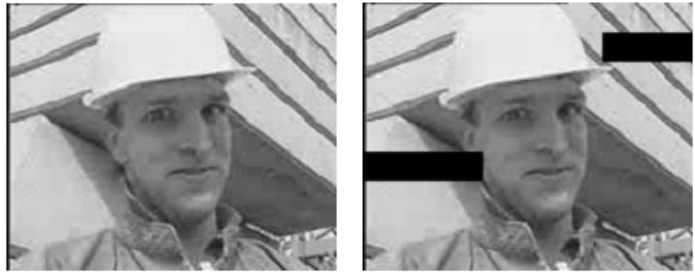

**Fig. 1.** (a) Decoded 14-th frame; and (b) Missing macro block location for the Foreman sequence (inter frame).

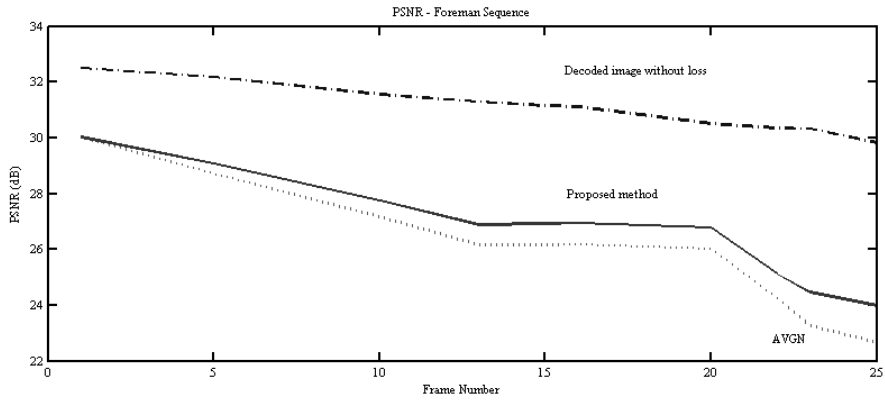

**Fig. 2.** PSNR plots for the Foreman sequence.

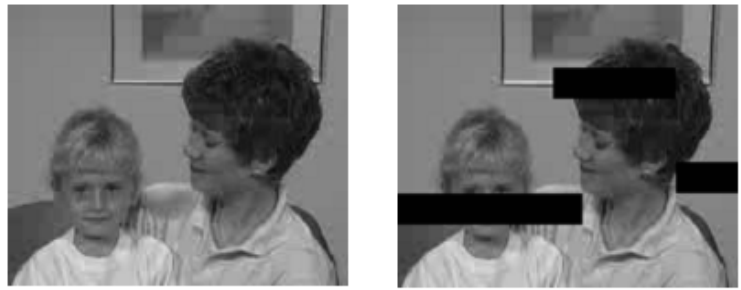

**Fig. 3.** (a) Decoded 8-th frame; and (b) Missing macro block location for the Mother and Daughter sequence (inter frame).

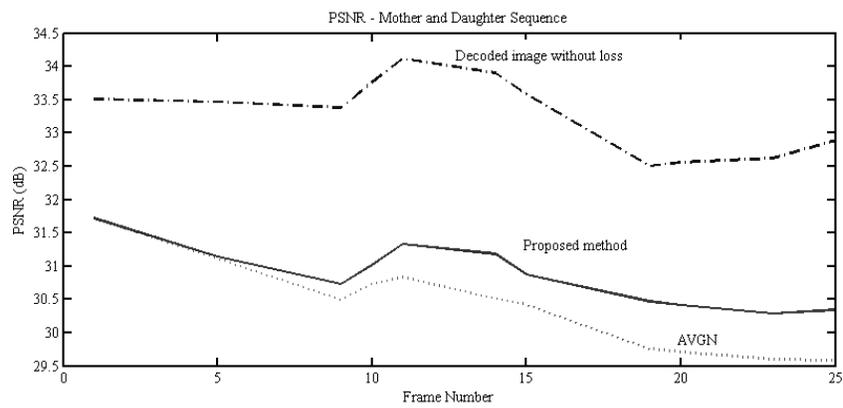

**Fig. 4.** PSNR plots for the Mother and Daughter sequence

## 4 Error Concealment

*A priori* information on the images as well as redundancies in both space (*h-v* directions) and time help to detect and correct errors. Intra frames of compressed video are basic frames that help generating inter frames: if intra frames contain lost data, then this will affect inter frames. This work assumes that missing MVs are correlated to the MVs of their neighbors. The proposed method assumes that a finite number of MV clusters exists. Each cluster corresponds to the displacement of a given region inside a frame (cf. [9, 13]). Once the received information is depacketized, a frame will have its pixels labeled as containing legitimate and erroneous regions. The legitimate ones can be used to calculate MVs with a simple procedure such as the one from Eq. (3). The motion recuperation algorithm from the previous section will be applied to cluster and correct the whole OF. The regularization operator

$$G(\hat{x}) = \frac{1}{1+q} \sum_{i=0}^{M-1} \sum_{j=0}^{M-1} \left\{ \hat{x}_{i,j} \left[ \frac{(\hat{x}_{i,j})^q - (\overline{x}_{i,j})^q}{q} \right] - (\overline{x}_{i,j})^q (\hat{x}_{i,j} - \overline{x}_{i,j}) \right\}$$

from Eq. (4) is critical since it incorporates prior knowledge about the original uncorrupted frame into the recovery problem. The choice of the regularization parameter α also affects the final result.

## 5 Experimental Results

Two *176×144* QCIF sequences were used: the "Foreman" and "The Mother and Daughter". The peak signal to noise ratio (*PSNR*) was chosen as a measure of performance. For an *8* bit *M×N* image, the *PSNR* in dB is given by

$$PSNR = 10 \log \left\{ \frac{255^2 MN}{\|w - \hat{w}\|^2} \right\},$$

where $w$ and $\hat{w}$ are, correspondingly, the original and restored images. The experiments considered a 5% cell loss. Fig. 1 and Fig. 3 show examples with a complete and a corrupted frame (i.e., missing macroblocks) for each sequence used. For all experiments, it was assumed that $q=1$ and $\gamma=0.8$. The algorithm was applied to compute $Q(x)$, by means of a constant regularization parameter α and with an optimal α. However, a constant α may yield bad convergence as the maximum number of iterations grows because the curve for $Q(x) \times \alpha$ resumes increasing after a few iterations. To some extent, increasing the value of $\alpha_0$ remediates this behavior. This becomes necessary due to the need to obey other bounds on errors and trends used to assess and to make a decision upon the optimum estimate selection. Multicriteria and an adaptive Bregaminized regularization algorithm give better results, although increasing the computational load. Experiments were also performed with a simple EC algorithm relying on the average of the MVs from neighbors (AVGN) and without loss of cells. Fig. 2 and Fig. 4 show *PSNR* plots when the three algorithms are applied to the

"Foreman" and the "Mother and Daughter" sequences. It is important to mention that the use of Bregman Divergence s improved the displacement vectors around borders.

## 6 Conclusions

This work solves the error concealment problem by means of Bregman divergence s and regularization [5,8,9,10], where the regularization function does not have to be fixed at each interaction any longer. It is important to devise lower and upper values for $\alpha^0$, so that convergence and optimization are more efficient. Certainly, the convergence, stability and merit of the estimates are improved when judged against to iterations relying on invariable values of $\alpha$. The fact that more local information on the image neighborhood is added to the regularization procedure is the main motive for these improvements. Occasionally, Bayesian estimation introduces estimation bias by prior information that may be needless. A possibility is to model estimation as a minimization of an expected Bregman divergence between the unknown and the projected distributions. It has been proven that Bregman iteration also approaches a solution with much less computational complexity than conventional regularization. Furthermore, when stopped according to the discrepancy principle, Bregman iteration is also a choice method for solving compressed sensing problems and they are very suitable for parallel implementations due to its characteristics [14-16]. Bregman divergences [13, 14] can be employed as measures of nearness—these divergences are natural for learning low-rank kernels since they maintain rank as well as positive semi-definiteness. Special cases of the proposed framework yield faster algorithms for several learning problems, and experimental results show that this algorithm can effectively learn both low-rank and full-rank kernel matrices.

## References

1. Bregman, M.: The Relaxation Method of Finding the Common Point of Convex Sets and its Application to the Solution of Problems in Convex Programming, USSR Comt. Math and Math. Ph., vol7, n.3, pp. 200-217, (1967).
2. Tikhonov, A.N., Arsenin, V.Y.: Solutions of Ill-Posed Problems, J. Willey & Sons, (1977).
3. Stutz, D.: Restauração de Imagens em Escala Nanométrica com Funcional de Regularização de Tikhonov e Computação Paralela, M.Sc. thesis, IPRJ/UERJ, N. Friburgo, RJ, Brazil, (2004).
4. Stutz, D., Silva Neto, A.J., Farias, R.C.: Information Weighted Mean Square Error (IWMSE): Uma Medida de Comparação de Imagens Baseada na Percepção, X EMC, N. Friburgo, RJ, Brazil, (2007).
5. Galatsanos, N.P., Katsaggelos, A.K.: Methods for Choosing the Regularization Parameter and Estimating the Noise Variance in Image Restoration and their Relation, IEEE Trans. on Im. Proc., pp.322-336, (1992).
6. Coelho, A. M. Estrela, V.V.: EM-Based Mixture Models Applied to Video Event Detection, In Principal Component Analysis - Engineering Applications, ISBN 9788563337214,

Intech, pp. 102-124, (2012).`http://www.intechopen.com/books/principal-component-analysis-engineering-applications/em-based-mixture-models-applied-to-video-event-detection`

7. Osher, S., Burger, M., Goldfarb, D., Xu, J.,Yin, W.: An Iterative Regularization Method for Total Variation Based Image Restoration, Multiscale Modeling Sim., 4, pp. 460–489, (2005).
8. N. Murata, T. Takenouchi, T. Kanamori, S. Eguchi, Information Geometry of U-Boost and Bregman Divergence, Neural Comput., vol.16, pp. 1437–1481, (2004).
9. Banerjee, A., Dhillon, I., Ghosh, J., Merugu, S.: An Information Theoretic Analysis of Maximum Likelihood Mixture Estimation for Exponential Families, Proc. 21st ICML, (2004).
10. Aubert, G., Kornprobst, P.: Mathematical Problems in Image Processing: Partial Differential Equations and the Calculus of Variations, 2nd ed., Springer, New York, (2006).
11. Hinterberger, W. Scherzer, O.: Models for Image Interpolation Based on the Optical Flow, Computing, 66, pp. 231–247, (2001).
12. H . Grossauer, Inpainting of Movies Using Optical Flow, Math. Models for Registration and Applications to Med. Imaging, Math. Ind. 10, Springer, Berlin, pp. 151–162, (2006).
13. Slesareva, N., Bruhn, A., Weickert, J.: Optic Flow Goes Stereo: a Variational Method for Estimating Discontinuity-Preserving Dense Disparity Maps, Proc. of the 27th DAGM Symp., Vienna, Austria, LNCS 3663, Springer, pp. 33–40, (2005).
14. Li, X., Jackson, J.R., Katsaggelos, A.K., Mersereau, R.M.: Multiple Global Affine Motion Model for H.264 Video Coding with Low Bit Rate, Proc. SPIE VCIP, San Jose, CA, (2005).
15. Coelho, A.M., Estrela, V.V., de Assis, J.T.: Error Concealment by Means of Clustered Blockwise PCA, IEEE. Picture Coding Symposium, Chicago, IL, USA, (2009).
16. do Carmo, F.P. , Estrela, V.V., de Assis, J.T.: Estimating Motion with Principal Component Regression Strategies, Proc. of IEEE MMSP '09), Rio de Janeiro, RJ, Brazil, (2009).
17. Coelho, A. M. Estrela, V. V.: Data-Driven Motion Estimation with Spatial Adaptation, Intl. J. of Image Proc. (IJIP), Vol. 6, Issue 1, pp. 53-67, (2012). `http://www.cscjournals.org/csc/manuscript/Journals/IJIP/volume6/Issue1/IJIP-513.pdf`